\documentclass{article}
\usepackage{graphicx} 
\usepackage{amsmath} 
\usepackage{hyperref} 
\usepackage{cleveref} 
\usepackage{algorithm}
\usepackage{algorithmic}
\usepackage{natbib}
\usepackage{mathtools}
\usepackage{amsfonts}
\usepackage{multirow} 
\usepackage{authblk}
\usepackage{xcolor}
\usepackage{subcaption}
\usepackage{subcaption}
\usepackage{booktabs}
\usepackage{multirow}
\usepackage{caption} 
\usepackage{appendix}
\usepackage{bm}
\usepackage[normalem]{ulem}

\title{Explainable Federated Bayesian Causal Inference and Its Application in Advanced Manufacturing}
\author[1]{Xiaofeng Xiao}
\author[2]{Khawlah Alharbi}
\author[2]{Pengyu Zhang}
\author[2]{Hantang Qin}
\author[1]{Xubo Yue\thanks{Corresponding Author: x.yue@northeastern.edu}}
\affil[1]{Mechanical \& Industrial Engineering, Northeastern University}
\affil[2]{Industrial and Systems Engineering, University of Wisconsin-Madison}

\begin{document}

\maketitle

\begin{abstract}
Causal inference has recently gained notable attention across various fields like biology, healthcare, and environmental science, especially within explainable artificial intelligence (xAI) systems, for uncovering the causal relationships among multiple variables and outcomes. Yet, it has not been fully recognized and deployed in the manufacturing systems. In this paper, we introduce an explainable, scalable, and flexible federated Bayesian learning framework, \texttt{xFBCI}, designed to explore causality through treatment effect estimation in distributed manufacturing systems. By leveraging federated Bayesian learning, we efficiently estimate posterior of local parameters to derive the propensity score for each client without accessing local private data. These scores are then used to estimate the treatment effect using propensity score matching (PSM). Through simulations on various datasets and a real-world Electrohydrodynamic (EHD) printing data, we demonstrate that our approach outperforms standard Bayesian causal inference methods and several state-of-the-art federated learning benchmarks.

\end{abstract}

\section{Introduction}

The increasing reliance on distributed data across organizations has led to a pressing demand for privacy-preserving methods. For example, hospitals are required to protect patient privacy and cannot share data without proper authorization. Similarly, autonomous driving systems are required to protect the private information of individual vehicles.  Importantly, even if all the data were allowed to be aggregated, the sheer volume would pose significant challenges for data storage and communication. Fortunately, federated learning has been developed to address these challenges (\cite{mcmahan2017communication}, \cite{jordan2019communication}), allowing decentralized clients to collaboratively train models while keeping their data local and only sending model information to a central server. Subsequently, the federated technique has been extended to various domains. To name a few, \cite{kontar2021internet} described its application prospects in the Internet of Things (IoT). \cite{xu2021federated} demonstrated the potential of applying federated learning in healthcare informatics. \cite{ghimire2022recent} explored the role of federated learning in enhancing cybersecurity. 

While many machine learning methods perform well in exploring correlations among variables in datasets, they often fall short in revealing intrinsic causality among variables.
Causality identifies the cause-and-effect relationships between variables, often determined through external interventions or controlled experiments \citep{pearl2010causal}. For instance, the treatment of a disease can result from taking medicine or from other interventions. Causal analysis can help determine the effect of each intervention independently, ultimately concluding the true effect of the medicine. Typically, estimating causal effect (treatment effect) is adopted to explore these relationships after interventions \citep{rubin1997estimating,hernan2006estimating}. This process of studying cause-and-effect relationships is known as \textbf{causal inference}. Furthermore, quantifying the uncertainty in these relationships is crucial to enhancing the reliability of the estimated causal effects. To address this, several studies, such as \cite{toth2022active} and \cite{li2023bayesian}, have introduced the Bayesian framework for computing treatment effect in causal inference. The Bayesian approach provides valuable uncertainty quantification for parameter estimation, ensuring more robust and interpretable results.

However, as data continues to grow and the systems that need to be processed become increasingly complex, there is a growing demand for efficiently and accurately exploring causality under data privacy in distributed datasets. Therefore, it is essential to explore methods for analyzing Bayesian causal inference in federated setting. A few existing studies have explored this approach. For instance, \cite{vo2021federated} designed the methodologies for analyzing causal effect from incomplete data in federated learning. \cite{xiong2023federated} estimated federated treatment effect using restricted  maximum likelihood estimator and applied them to heterogeneous observational medical data. 
However, the limitations of these studies in terms of scalability are evident, and the implementation of these methods remains relatively complex. In addition, we observed that the application of current federated Bayesian causal inference approaches is still quite narrow. Especially, fewer research has explored its application in advanced manufacturing. Causal inference plays a crucial role in explainable artificial intelligence (xAI) systems \citep{chou2022counterfactuals} which serves as the foundation for advanced manufacturing \citep{ahmed2022artificial}. Meanwhile, the requirements for data privacy rules and massive data storage are becoming increasingly stringent. Consequently, the substantial influence of federated Bayesian causal inference on advanced manufacturing must be recognized.


To address the above concerns and limitations, this paper proposes an approach of explainable Bayesian federated causal inference (\texttt{xFBCI}) to explore the treatment effect in manufacturing systems. This framework consists of two phases: (1) We develop a Bayesian federated variational inference framework to estimate the personalized posterior of each client's parameters. Specifically, we employ the Expectation Propagation (\texttt{EP}) to approximate the posteriors of clients from the global server without accessing to local data. The Bayesian framework enhances the explainability of our method, and the \texttt{EP} algorithm improves its scalability.  
(2) After obtaining the personalized parameters of each client, we adopt propensity score matching (PSM) \citep{rubin1997estimating} to derive the treatment effect of each covariate in the client's dataset. PSM has the advantage of mapping all covariates as propensity scores, effectively addressing the confounding effects among these variables. Overall, the proposed xFBCI allows us to easily gather causal effects among variables in each client without data sharing within the manufacturing system. To demonstrate the functionality of \texttt{xFBCI}, we apply our approach to Electrohydrodynamic (EHD) printing, a key area in additive manufacturing.

\subsection{Motivation}

Our work is motivated by the pressing need to address these challenges in EHD applications. EHD printing has emerged as a prominent additive manufacturing technique, celebrated for its ability to create intricate micro- and nanoscale patterns and structures. Electrostatic forces are utilized to drive a conductive solution through an extrusion nozzle, generating a fine jet or droplets that are precisely deposited onto the substrate, as the schematic that shown in Figure 1. EHD printing has diverse applications across multiple manufacturing fields, such as electronics, biomedicine, and aerospace \citep{wolf2024heterogeneous, JIANG2024248}. Several variables affect the quality and precision of the printed structure, such as voltage, nozzle moving speed, and nozzle size. Understanding the causality between these variables and printing outcomes is crucial for improving the performance of EHD printers. At the same time, these variables can influence each other, leading to confounding effects. Therefore, addressing the confounding effect among the variables is necessary. Nowadays, multiple EHD printers may operate at different locations as part of a system to complete complex additive manufacturing tasks. These printers use the same technology but in different situations or for different purposes. For example, EHD printers may work on the ground environment as well as in a zero-gravity environment for in-space manufacturing applications \citep{jiang2024situ}. The causality between them can be analyzed and provide useful insights for future work. However, the dataset from each EHD printer in this system can be enormous, making communication difficult. More importantly, these datasets are often required to remain confidential and cannot be shared. As a result, learning the causal effects of variables, communicating datasets efficiently, and protecting data privacy simultaneously would significantly improve the performance of the EHD printing system. However, to our knowledge, there are few to no prior studies addressing the above-mentioned issues related to EHD printing. For this reason, we utilize \texttt{xFBCI} to a realistic EHD printing dataset as a proof of concept. Furthermore, our results highlight the effectiveness of our approach (See Secs. \ref{sec:simulation} and \ref{sec:EHD}), and we are optimistic about its potential to be generalized across a broader range of manufacturing domains.

\begin{figure}
    \centering
    \includegraphics[width=0.8\linewidth]{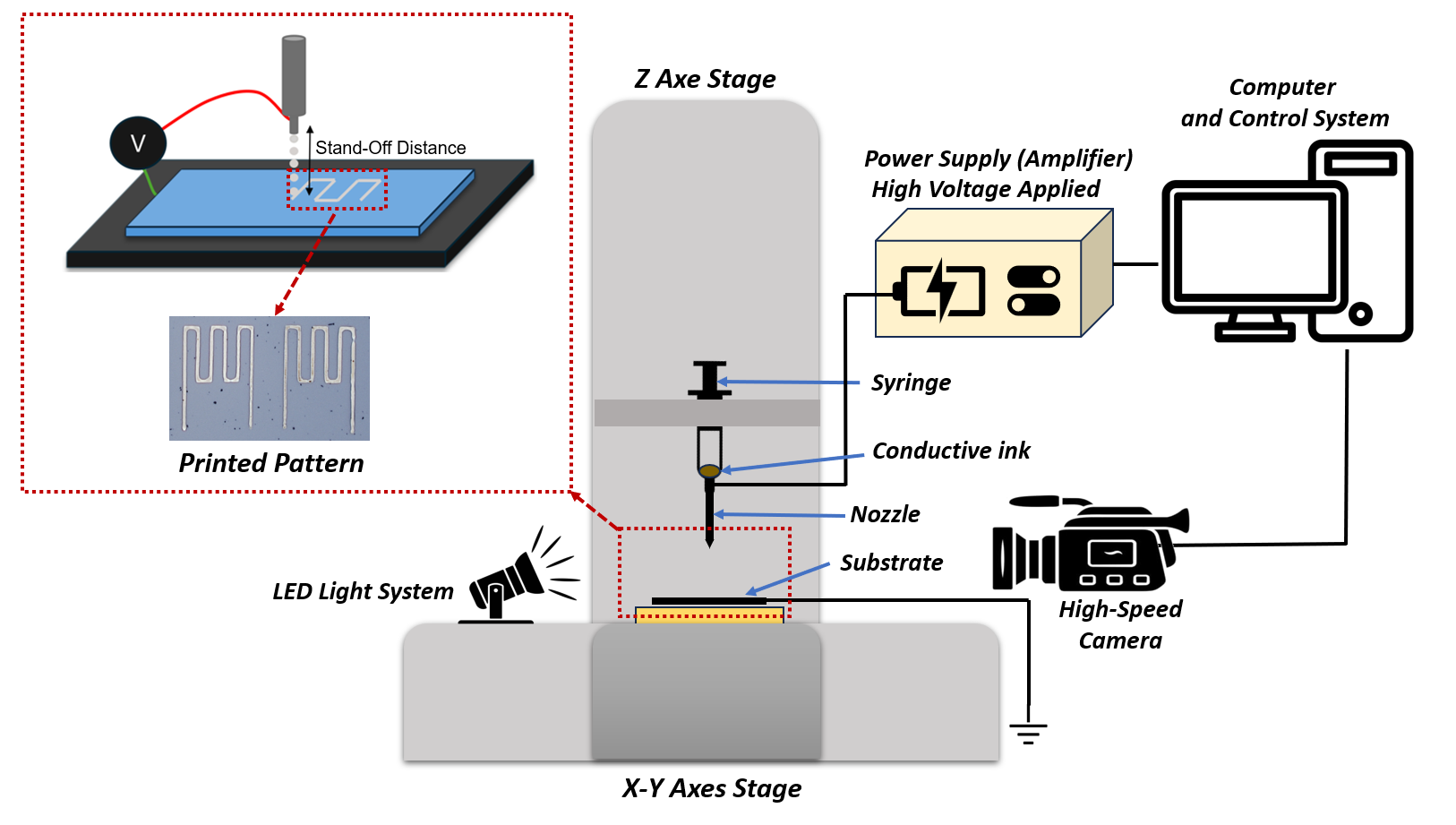}
    \caption{Schematic of EHD printing setup.}
    \label{fig:enter-label}
\end{figure}




\subsection{Contributions}

The contributions of this paper are described below: 

\begin{itemize}
    \item We propose \texttt{xFBCI}, an explainable Bayesian federated approach for causal inference. With \texttt{EP}, our framework achieves advanced scalability and simplified implementation. Notably, we use FL solely to estimate propensity scores and perform matching. The subsequent predictive models applied to the matched data are model-agnostic, simplifying the process by eliminating the need for separate likelihood construction for each client.
    \item By leveraging propensity score methods to estimate the Average Treatment Effect and applying matching to enhance predictive accuracy, we provide a robust framework for understanding causal relationships and improving predictive modeling in optimizing EHD printing processes. This proposed method alleviates the gap in existing studies on federated causal inference in advanced manufacturing. 
    \item \texttt{xFBCI} functions successfully in EHD printing, aiding in the design of EHD printers with improved printing quality.
\end{itemize}
Our fully annotated Python code is available on Github: \href{https://github.com/xx987/xFBCI/tree/main}{\texttt{xFBCIcode}}.

The remainder of this paper is organized as follows: In Section \ref{sec:review}, we review some related works on Bayesian federated learning and causal inference. Then, we present the details of our proposed \texttt{xFBCI} in Section \ref{sec:method}. The results of simulations are displayed in Section \ref{sec:simulation}. Finally, we present the outcomes of \texttt{xFBCI}'s application in EHD printing in Section \ref{sec:EHD}. We conclude this paper with a brief discussion in Section \ref{sec:con}.

\section{Literature Review} \label{sec:review}
In this section, we provide a concise review of related work on Bayesian federated learning and causal inference.

\paragraph{Bayesian Federated Learning.} Federated learning is a machine learning paradigm that enables multiple clients to collaboratively train a shared model while keeping their data locally. The seminal paper \citep{mcmahan2017communication} proposed a simple and elegant federated averaging (\texttt{FedAvg}) algorithm. Specifically, each client performs local stochastic gradient descent (SGD) and then sends local model parameters to a global server for aggregation. The global server takes a weighted average to aggregate the local parameters. The aggregated global parameter is then sent back to the local client for optimization. This iterative communication process continues until convergence or some exit condition is met. Since then, there has been a significant surge in research papers on FL. Please refer to \citep{li2020review, kontar2021internet} for comprehensive reviews of the applications of FL in various Engineering domains. To date, compared to the vast body of existing FL research, there have been relatively few papers focused on Bayesian FL \citep{cao2023bayesian}.  Notably, \cite{al2020federated} introduced Federated Posterior Averaging (\texttt{FedPA}), a method enabling clients to efficiently communicate with the server by sharing the posterior distributions of parameters. \cite{chen2020fedbe} adopted Bayesian model ensemble strategy (\texttt{FedBE}) to aggregate the global model efficiently. \cite{kotelevskii2022fedpop} introduced a novel stochastic optimization method based on Markov Chain Monte Carlo (MCMC) for personalized federated learning. This approach utilizes a Bayesian framework to effectively address uncertainty in parameter estimation. Due to space limitation, please refer to \cite{cao2023bayesian} for a comprehensive review. 


\paragraph{Bayesian Causal Inference.} One of the primary objectives in causal inference is estimating the treatment effect. A seminal work by \citet{rubin2005causal} introduced the potential outcomes framework, providing a foundation for computing treatment effects under both randomized experiments and observational studies. To model outcomes with greater accuracy and flexibility, \citet{hill2011bayesian} proposed a Bayesian nonparametric approach known as Bayesian Additive Regression Trees (BART). This method constructs tree-based models for observed data and outcomes, effectively capturing complex relationships, including nonlinearity. More recently, \citet{li2023bayesian} reviewed various Bayesian strategies for causal inference, including a novel perspective that frames potential outcomes as a missing data problem. In this approach, Bayesian methods are applied to sample missing outcomes, enabling robust estimation of causal effects.


Exploring causal inference within FL setting is critical, as there is a growing demand for interpretable artificial intelligence and privacy. However, only a few studies have examined causal inference in a federated approach. For instance,   
\cite{vo2021federated} leveraged Gaussian Process to handle outcome model imputation and then decomposed this model across clients to estimate the causal effect from incomplete data in federated learning. Subsequently, \cite{vo2022adaptive} extended the approach of Random Fourier Features to estimate the causal effect (\texttt{CausalRFF}). More recently, \cite{xiong2023federated} used the propensity scores to obtain federated statistics estimators of treatment effects. \cite{kayaalp2024causal} examined a setting where heterogeneous agents with connectivity perform inference using unlabeled streaming data.

\paragraph{Causal Inference in Manufacturing.} Causal inference plays a crucial role in advanced manufacturing, as it serves as the foundation for explainable artificial intelligence (xAI) systems. These systems, in turn, enable manufacturing processes to become more sophisticated by providing clear insights into the causal relationships driving system behaviors. For instance, \cite{chen2021ontology} proposed a Bayesian network (BN) model to represent causal relationships in additive manufacturing (AM), thereby enhancing AM's interoperability and controllability. Similarly, \cite{oliveira2021understanding} applied causal inference to perform true root cause analysis in manufacturing. Moreover, \cite{hua2022zero} presented a framework that combines convolutional neural networks and causal inference to predict key components in intelligent manufacturing.



\section{Method}
\label{sec:method}
In this section, we detail our proposed \texttt{xFBCI} framework for scalable and privacy-preserving propensity score matching in causal inference. In Sec.~\ref{sec:nota}, we introduce the problem settings and notations. Sec.~\ref{title:1} outlines the federated Bayesian inference algorithm. The procedures for estimating treatment effects are detailed in Sec.~\ref{sec:TE}. Our approach allows each local client to collaboratively learn a global posterior without data sharing. The learned posterior is then personalized for each client to estimate propensity scores and subsequently estimate average treatment effect. This approach facilitates knowledge sharing to enhance treatment effect estimation in a privacy-preserving setting. 


\subsection{Notations}\label{sec:nota}
We begin by introducing our problem setting to enhance clarity in the following sections. We consider a collaborative environment with $K \geq 2$ clients. For client $k \in[K]:=$ $\{1, \ldots, K\}$,  the corresponding dataset is $\mathcal{D}_k=\left\{X_k, W_k, Y_k\right\}$, comprising $N_k$ observations, where $Y_k=\left[y_{k 1}, \ldots, y_{k N_k}\right]^{\top}$ is a $N_k \times 1$ output vector, and $W_k=\left[w_{k 1}, \ldots, w_{k N_k}\right]^{\top}$ is a $N_k \times 1$ vector representing if the unit was treated. Without loss of generality, we consider binary $W_k$ here, where $w_{ki} = 1$ means the sample unit is treated. Otherwise $w_{ki} = 0$, $\forall i=1, \ldots, N_k$. Please note that our method can be easily extended to accommodate multi-class treatments. In $\mathcal{D}_k$, $X_k=\left[x_{k 1}, \ldots, x_{k N_k}\right]$ is a $d \times N_k$ input matrix, where $x_{k 1}=\left(\left[x_{k 1}\right]_1, \ldots,\left[x_{k 1}\right]_d\right)^{\boldsymbol{\top}}$, and $d$ represents the dimension of the input space. In the causal inference community, $Y_k$ is referred to as an outcome vector. The components $Y_k(0)$ and $Y_k(1)$, associated with $W_k$, represent the potential outcomes under control and treatment respectively, and are used to estimate the treatment effect (See details in Sec \ref{sec:TE}).


Following the work of \cite{rubin2005causal}, the estimated outcome could be derived upon the potential outcome model in the form of expected function: 
$$
\mathbb{E}[Y_{ki} \mid W_{ki}=0, x_{ki}] = \mathbb{E}[Y_{ki}(0) \mid x_{ki} ],
$$
$$
\mathbb{E}[Y_{ki} \mid W_{ki}=1, x_{ki}] = \mathbb{E}[Y_{ki}(1) \mid x_{ki}].
$$
Here we use capital letters (e.g., $Y_{ki}$) to denote random variables and lowercase letters (e.g., $y_{ki}$) to represent their corresponding realizations.
The total expectation of outcome can be written as: 
\begin{align*}
\mathbb{E}[Y_{ki}] &= \mathbb{E}[Y_{ki} \mid W_{ki}=1, x_{ki}] \cdot Pr(W_{ki}=1 \mid x_{ki}) \\
&\quad + \mathbb{E}[Y_{ki} \mid W_{ki}=0, x_{ki}] \cdot Pr(W_{ki}=0 \mid x_{ki}),
\end{align*} where  $Pr(W_{ki}=1 \mid x_{ki})$ represents the probability that the corresponding samples $x_{ki}$ is treated, and $Pr(W_{ki}=0 \mid x_{ki}) =1 - Pr(W_{ki}=1 \mid x_{ki}) $. Notably, $Pr(W_{ki}=1 \mid x_{ki})$ is also known as the \textbf{propensity score} in causal inference.


More formally, the propensity score is modeled as a probability: $$e\left(x_{ki}\right)  \coloneqq  Pr\left(W_{ki} = 1 \mid x_{ki}\right), \forall i=1, \ldots, N_k.$$ 

Typically, a logistic regression model (or any classification model) is used to estimate this probability:
$$
\log \left(\frac{Pr\left(W_{ki}=1 \mid x_{ki}\right)}{1-Pr\left(W_{ki}=1 \mid x_{ki}\right)}\right)=x_{ki}^{\top} \theta,
$$
where $\theta$  is the target parameter and $\theta \in \mathbb{R}^d$.

This logistic regression model motivates us to build up the likelihood for client $k$ as

\begin{equation}
p_k(\mathcal{D}_k \mid \theta)=\prod_{i=1}^{N_k}\left(\frac{1}{1+\exp \left(-x_{ki}^T \theta\right)}\right)^{w_{ki}} \cdot\left(\frac{\exp \left(-x_{ki}^T \theta\right)}{1+\exp \left(-x_{ki}^T \theta\right)}\right)^{1-w_{ki}}.\label{eq:1}
\end{equation}

Then, we consider the global posterior by employing the collection of client likelihoods: 

\begin{equation}
p(\theta|\mathcal{D})\propto p_0(\theta)\prod_{k=1}^K p_k({\mathcal{D}_k \mid \theta})\label{eq:glikeli},
\end{equation}
where $\mathcal{D}=\bigcup_{k \in[K]} \mathcal{D}_k$ is  the set of all clients' dataset, $p_0(\cdot)$ is a prior. We treat the prior distribution $p_0(\theta)$ of each client as flat prior for the fairness of sampling in each client.

\subsection{Federated Bayesian Variational Inference}\label{title:1}

Learning parameters from the joint posterior in Eq~\eqref{eq:glikeli} requires data sharing between clients and the server, which violates data privacy rules in the federated setting. Thus, we consider the federated Bayesian variational inference approach to tackle this challenge.

Variational inference approximates probability densities through optimization \citep{blei2017variational}. In our proposed approach \texttt{xFBCI}, we first introduce the approximated local likelihood $q_{k}(\theta)$ and approximated global posterior $q_{global}(\theta)$ for $p(\theta|\mathcal{D})$. Subsequently, Eq~\eqref{eq:glikeli} is approximated as:

\begin{equation}
q_{global}(\theta) \propto p_0(\theta)\prod_{k=1}^K q_{k}(\theta)\label{eq:likeli}.
\end{equation}

However, the federated setting only allows us to learn $q_{k}(\theta)$ locally. Specifically, each client cannot directly send $q_k(\theta)$ to the server as $q_k(\theta)$ will reveal $p_k({D_k \mid \theta})$. Therefore, we will leverage the expectation propagation (\texttt{EP}) \citep{minka2013expectation} to address this problem, and the details are provided below:

\textbf{(1)} We first calculate cavity distribution $q_{-k}({\theta}) \propto \frac{q_{{global }}({\theta})}{q_k({\theta})}$ to remove impact of $q_k({\theta})$ from global. \textbf{(2)}  Next, we compute the tilted distribution $q_{\backslash k}({\theta}) \propto p_k(\mathcal{D}_k \mid \theta) q_{-k}({\theta})$. This step incorporates the true $p_k(\mathcal{D}_k \mid \theta)$ into the cavity.  Importantly, the resulting product cannot reveal $p_k(\mathcal{D}_k \mid \theta)$ directly since it is merged with $q_{-k}({\theta})$. \textbf{(3)} Finally, we solve the optimization problem related to $q_{\backslash k}({\theta})$, deriving its estimator $\widehat{q}_{\backslash k}(\boldsymbol{\theta})$, and update $q_k^{\text{new}}({\theta})$ using $\Delta q_k=\widehat{q}_{\backslash\mathrm{k}} / q_{\text {global}}$.

Iteratively, we are filling the ``cavity" of global posterior by $p_k(\mathcal{D}_k\mid\theta)$. Ultimately, we will be able to approximate the true global posterior well. The step \textbf{(3)} in our \texttt{xFBCI} are summarized as an optimization problem below:

\begin{equation}
q_k^{\text {new }}({\theta})=\underset{q \in \mathcal{Q}}{\arg \min } D(\underbrace{p_k(\mathcal{D}_k \mid \theta) q_{-k}({\theta})}_{\propto q_{\backslash k}({\theta})} \| \underbrace{q({\theta}) q_{-k}({\theta})}_{\propto \widehat{q}_{\backslash k}({\theta})}) \label{eq:4}
\end{equation}
where $q_{-k}({\theta}) \propto \frac{q_{{global }}({\theta})}{q_k({\theta})}$, $\mathcal{Q}$ is the family of variational distribution, and $D(\cdot \| \cdot)$ is a divergence function (e.g., KL-divergence).

To enable efficient optimization, we treat the approximated posterior as Gaussian distributions $\mathcal{N}({\cdot},{\cdot})$. Specifically, the approximated local and global posterior are represented as: 
\begin{equation}
q_k({\theta})=\mathcal{N}({\theta} ; {\mu_k}, {\Sigma_k})=\mathcal{N}({\theta} ; {\eta_k}, {\Lambda_k}), \text { where } {\Lambda_k}:={\Sigma_k}^{-1}, {\eta_k}:={\Sigma_k}^{-1} {\mu_k} 
\end{equation}
\begin{equation}
q_{global}({\theta})=\mathcal{N}({\theta} ; {\mu_{g}}, {\Sigma_{g}})=\mathcal{N}({\theta} ; {\eta_{g}}, {\Lambda_{g}}), \text { where } {\Lambda_{g}}:={\Sigma_{g}}^{-1}, {\eta}:={\Sigma_{g}}^{-1} {\mu_{g}}. 
\end{equation}

The form of Gaussian distribution makes the expectation propagation be convenient. For instance, the production and quotation of Gaussian Distribution can computed as follow \citep{guo2023federated, yue2024federated}: 
\begin{equation}
\mathcal{N}\left(\theta ; \eta_1, \Lambda_1\right) \mathcal{N}\left(\theta ; \eta_2, \Lambda_2\right) \propto \mathcal{N}\left(\theta ; \eta_1+\eta_2, \Lambda_1+\Lambda_2\right),\label{eq:5}
\end{equation}

\begin{equation}
    \frac{\mathcal{N}\left(\theta ; \eta_1, \Lambda_1\right)}{\mathcal{N}\left(\theta ; \eta_2, \Lambda_2\right)} \propto \mathcal{N}\left(\theta ; \eta_1-\eta_2, \Lambda_1-\Lambda_2\right).\label{eq:6}
\end{equation}

Despite employing the Gaussian approximation, directly solving Eq~\eqref{eq:4} is still computational challenging, especially when $p_k(\mathcal{D}_k|\theta)$ does not have closed form. Therefore, we utilize Stochastic gradient Langevin dynamics (SGLD) \citep{welling2011bayesian} to approximate the $q_{\backslash k}({\theta})$. SGLD will make Eq~\eqref{eq:4} as an efficient stochastic optimization problem. Specifically, the $q_{\backslash k}({\theta})$ could be transferred to the amiable logarithm formation: 
\begin{equation}
\log q_{\backslash k}({\theta})  = \log p_k(\mathcal{D}_k \mid \theta) + \log q_{-k}({\theta}).
\end{equation}

Then, the SGLD samples $\theta$ by updating:

\begin{equation}
\Delta \theta_t=\frac{\varepsilon_t}{2}\left(\frac{N_k}{n}\nabla p_k(\mathcal{D}_k \mid \theta_t)+ \nabla \log q_{-k}({\theta_t})\right)+\phi_t\label{eq:8},
\end{equation}

\begin{equation}
\theta^{new}_t= \theta_t + \Delta \theta_t  ,
\end{equation}
where $t$ is the $t^{th}$ step of this iteration, $n$ is the batch size of each iteration, $N_k$ is a positive integer representing the number of observations (data points), $\phi_t \sim \mathcal{N}\left(0, \varepsilon_t\right)$ is Gaussian noise, and $\sum_{t=1}^{\infty} \varepsilon_t=\infty, \quad \sum_{t=1}^{\infty} \varepsilon_t^2<\infty$. We note that expectation propagation may produce a negative covariance matrix during iterative inference, preventing the local and global approximations from being positive definite. To address this problem, we follow the approach from \cite{vehtari2020expectation}, which involves performing eigendecomposition of the negative covariance matrix and then replacing all of the negative eigenvalues with small positive real numbers. This restructuring ensures that the covariance matrix remains positive definite throughout inference without causing significant deviations in the estimated values from the original approximation.

We collect the updated $\theta^{new}_t$ in each step of SGLD. After completing the iterations of SGLD, we obtain the collected set $\Theta=\left\{\theta_1, \theta_2, \ldots\right\}$, where $\theta_1, \theta_2, \ldots$ are $\theta^{new}_t$ at each iteration.

Intuitively, $q_{-k}(\theta)$ can be roughly viewed as a prior. Combining $q_{-k}(\theta)$ with $p_k(\mathcal{D}_k \mid \theta)$, we can find that SGLD efficiently samples the posterior $\theta$ from $p_k(\mathcal{D}_k \mid \theta)q_{-k}(\theta)$. Finally, we compute the average of the collected $\Theta$, denoted as  $\hat{\theta}_t$. This $\hat{\theta}_t$ serves as the first moment estimator of $\theta$ in $q_{\backslash k}({\theta})$, and the covariance is estimated by using the scaled identity matrix $\alpha I$ \citep{guo2023federated}, where $\alpha$ is a hyperparameter and $I$ is an identity matrix  .

Notably, in the \texttt{xFBCI} framework, the likelihood $p_k(\mathcal{D}_k \mid \theta)$ of all clients' datasets are mapped to the form of \eqref{eq:1}, which is a logistics format. This advantage of our \texttt{xFBCI} significantly reduces computational complexity, as we do not need to construct a likelihood function for the dataset of each client separately. Furthermore, the subsequent predictive models applied to the dataset are model-agnostic, simplifying the process by eliminating the need for separate likelihood construction for each client.

\subsection{Treatment Effect}\label{sec:TE}

The updated $q_k^{\text {new}}({\theta})$ from the previous section contains information about the parameter $\theta$ from both global and local information. This personalized parameter will be used to determine the propensity score for each client, thereby deriving the treatment effect of the covariates. The propensity score of each client is modeled as a sigmoid function representing the probability, and we estimate it by applying $\hat{\theta}_k$ (e.g., mean of the density $q_k^{\text {new }}({\theta})$) as

\begin{equation}
Pr(W_{ki}=1 \mid x_{ki})=
\frac{\exp \left(x_{ki}^{\top} \hat{\theta}_k\right)}{1+\exp \left(x_{ki}^{\top} \hat{\theta}_k\right)}\label{eq:9}.
\end{equation}

After obtaining the propensity scores for all samples in the client, we explore the causality of covariates in each client by calculating the treatment effect. One of the common approaches is matching the samples with close propensity scores. 
To ensure scalability, we perform the Nearest neighbor matching (NNM): 

\begin{equation}j=\arg \min _{j: w_{kj}=0}\left|Pr_m-Pr_j\right|,
\end{equation}
where $Pr_m$ is the propensity score of the treated unit $m$, and $Pr_j$ is the propensity score of the control unit $j$. NNM matches propensity scores by minimizing the difference between treated and control groups.

Finally, we estimate the average treatment effect (ATE) of the potential outcome $Y_i$ when treated or controlled:
$$
\hat{\tau}:=\mathrm{E}\left[Y_{ki}(1)-Y_{ki}(0)\right].
$$
Our algorithm is summarized as follow: 
\begin{algorithm}[H]
\caption{explainable Federated Bayesian Causal Inference \label{eq:a1}}
\begin{algorithmic}[1] 
    \STATE \textbf{Input:} Data $\{X_k, W_k, Y_k\}_k$, number of clients $K$, number of communication rounds $T$, initial approximation $\left\{q_k(\theta)\right\}_{k=1}^K$, flat prior $p_0(\theta)$,  $q_{global}(\theta) = p_0(\theta)\prod_{k=1}^K q_{k}(\theta)$, $\delta = 1/K$ 
    \STATE \textbf{Output:} Estimated Average Treatment Effect: $\hat{\tau}$
    \STATE Initialize parameters in $\theta$
    \FOR{$i = 1$ to $T$}
        \STATE Broadcast $q_{global}(\theta)$
        \FOR{$k \in K$} 
            \STATE Compute $q_{-k}({\theta})$ using $\frac{q_{{global }}({\theta})}{q_k({\theta})}$
            \STATE Compute estimator $\widehat{q}_{\backslash k}({\theta})$ of $q_{\backslash k}({\theta})\propto p_k(\mathcal{D}_k \mid \theta) q_{-k}({\theta})$ using SGLD as Eq~\eqref{eq:8}
            \STATE Compute $\Delta q_k({\theta}) = \widehat{q}_{\backslash k}({\theta})/ q_{global}{({\theta})}$ 
        \ENDFOR
        \STATE Update\\
        Client: $q_k^{\text {new }}(\theta) \propto q_k(\theta)\left(\Delta q_k(\theta)\right)^\delta \quad$\\
        Server: $q_{global}^{\text {new}}(\theta) \propto q_{global}(\theta) \prod_k\left(\Delta q_k(\theta)\right)^\delta$
        
    \ENDFOR
    \STATE Collect $\hat{\theta}_k$ from $q_{k}^{\text {new}}(\theta) $
     to compute $\hat{\tau}$ in each client as shown in Sec.~\ref{sec:TE}
    \STATE \textbf{return} $\hat{\tau}$
\end{algorithmic}
\end{algorithm}

\section{Simulation} \label{sec:simulation}

In this section, we present a series of simulation studies to evaluate the performance of \texttt{xFBCI}. These studies illustrate \texttt{xFBCI}'s ability to deliver accurate parameter estimates and precise ATE calculations, even under various challenging conditions. Our proposed \texttt{xFBCI} model will be evaluated against the following benchmarks: (1) \texttt{Individual}: Each local client will independently perform optimization without any communication with other clients; (2) \texttt{FedAvg} with personalization technique \texttt{Ditto}: \texttt{Ditto} first computes a global model parameter using \texttt{FedAvg}. Then, each client solves a constrained optimization problem to retain information from the global parameter of \texttt{FedAvg} to obtain its personalized parameter (\cite{li2021ditto}). \texttt{Ditto} is currently one of the state-of-the-art personalized FL algorithms. (3) \texttt{Central}: Data from all clients are combined to perform optimization and calculate the ATE. 

 

\subsection{One Simple Homogeneous Case}

\paragraph{Case 1.} We start with a simple experiment using exactly the same parameters across all of clients. For each client, we simulate the dataset following the \cite{vo2021federated} as follows: 
$$
\begin{aligned}
X_k & \sim \mathrm{U}(-1,1), & Y_k(0) \sim \mathcal{N}\left(\lambda\left(b_0+X_k^{\top} \mathbf{b}_1\right), \sigma_0^2\right), \\
W_k & \sim \operatorname{Bern}\left(\varphi\left(a_0+X_k^{\top} \mathbf{a}_1\right)\right), & Y_k(1) \sim \mathcal{N}\left(\lambda\left(c_0+X_k^{\top} \mathbf{c}_1\right), \sigma_1^2\right),
\end{aligned}
$$where $\varphi(\cdot)$ denotes the sigmoid function, $\lambda(\cdot)$ denotes the softplus function, and $X_k=\left[x_{k 1}, \ldots, x_{k N_k}\right]$ is a $d \times N_k$ matrix with $d=5$, $x_{k 1}=\left(\left[x_{k 1}\right]_1, \ldots,\left[x_{k 1}\right]_d\right)^{\boldsymbol{\top}}$, and $Y_k=\left[y_{k 1}, \ldots, y_{k N_k}\right]^{\top}$, and $W_k=\left[w_{k 1}, \ldots, w_{k N_k}\right]^{\top}$ is a $N_k \times 1$ vector representing if the unit was treated. The components $Y_k(0)$ and $Y_k(1)$, associated with $W_k$, represent the potential outcomes under control and treatment respectively.


The true parameters are set as follows: $\sigma_0=\sigma_1=0, \left(a_0, b_0, c_0\right)=(0.6,6,30)$, ($\mathbf{a}_1, \mathbf{b}_1, \mathbf{c}_1) = (\boldsymbol{0}, \boldsymbol{10}, \boldsymbol{15})$, where $\boldsymbol{0}=(0, 0, 0, 0, 0)^\top$, $\boldsymbol{10}=(10, 10, 10, 10, 10)^\top$, $\boldsymbol{15}=(15, 15, 15, 15, 15)^\top$. 

We generate datasets for 5 clients, denoted as $k \in\{1, \ldots, 5\}$, with each client having 1000 samples (balanced datasets).

\paragraph{Evaluation Metrics.} The evaluation metrics include the average root mean square error (A-RMSE) for the parameters, the average estimated ATE (A-ATE), and the deviation of the estimated ATE from the true ATE. A-RMSE is defined as
$$
\text{A-RMSE} = \frac{1}{K} \sum_{k=1}^K \sqrt{\frac{\sum_{i=1}^{d}\left(\hat{\theta}_k^i - {\theta}_k^i\right)^2}{d}},
$$where ${\theta}_k$ and $\hat{\theta}_k$ are the real and estimated parameters from the simulated dataset in client $k$.
Similarly, the A-ATE is defined as: 
$$
\text{A-ATE} = \frac{1}{K} \sum_{k=1}^K \tau_k.
$$where $\tau_k$ is the estimated ATE of client $k$.

We repeat the above simulation process for 10 replications. The mean values (and the standard errors in the bracket) for each baseline of the evaluation metrics are reported in Table~\ref{tab:table 1}. Specifically, the ATE error is calculated as the absolute value of mean difference between each replication’s A-ATE and the true ATE. 

It can be seen that our proposed \texttt{xFBCI} outperforms the \texttt{individual} and \texttt{Ditto} in terms of both parameter and ATE estimations. Understandably, the centralized benchmark outperforms the other three benchmarks. By aggregating data from all clients, the centralized approach captures the true patterns of the entire dataset, resulting in a more accurate estimation. However, this approach violates the data privacy rules in federated learning.

\begin{figure}[h]
    \centering
    \captionsetup{font=footnotesize}
    \begin{subfigure}[b]{0.45\textwidth} 
        \centering
        \includegraphics[width=\textwidth]{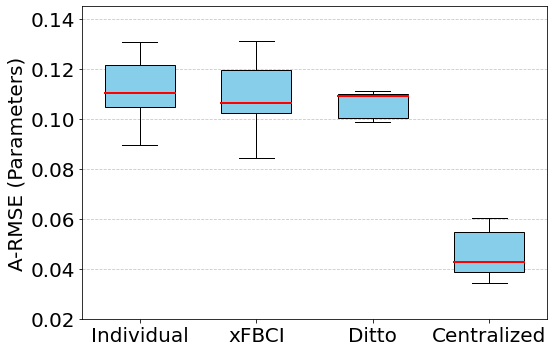}
        \caption{A-RMSE}
        \label{fig:dd}
    \end{subfigure}
    \hfill 
    \begin{subfigure}[b]{0.45\textwidth}
        \centering
        \includegraphics[width=\textwidth]{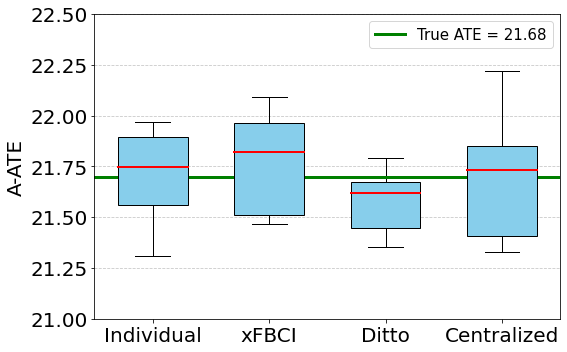}
        \caption{A-ATE}
        \label{fig:ee}
    \end{subfigure}
    \caption{Boxplot of the results from 10 replications in Case 1. In (b), the green line indicates the mean value of the true ATE across 10 replications.}
    \label{fig:figure 1}
\end{figure}

We present the boxplot of all results in Figure~\ref{fig:figure 1}. Specifically, the mean value of the true ATE over 10 replications is shown in Figure~\ref{fig:figure 1}(b) as the \textcolor{green}{green line}. The figure illustrates that \texttt{xFBCI} achieves a reliable estimation of the ATE even in a federated setting.

\begin{table}[H]
    \centering
\captionsetup{font=footnotesize}
\begin{tabular}{c c c c}
\hline Models & A-RMSE & A-ATE & Error of ATE \\
\hline \texttt{Individual} & 0.106($\pm 0.012$)  & 21.73($\pm 0.15$) &0.05 \\
\hline \textbf{\texttt{xFBCI}} & \bm{$0.103(\pm 0.010)$} &21.70($\pm 0.14$) &  \bm{$0.02$} \\
\hline \texttt{Ditto}  &0.112($\pm0.004$) &21.58($\pm0.07$) & 0.10 \\
\hline \texttt{Centralized} &0.051($\pm 0.019$) &21.69($\pm 0.17$) &0.01 \\
\hline

\end{tabular}
    \caption{Results of Case 1. We report the mean values and standard errors of 10 replications for each metric across the 4 benchmarks. The sample size of each client is: $N_k = 1000$, where $k \in\{1, \ldots, 5\}$.}
\label{tab:table 1}
\end{table}

\subsection{Heterogeneous Cases}

Unlike the previous section, the heterogeneous settings involve adjusting the dimensions, sample sizes, and true data-generating parameters according to the cases discussed herein.

\paragraph{Case 2.} In this experiment, the dataset generation process is similar to \textbf{Case 1}, except that the true parameters are sampled randomly. In case 2, the true parameters are set as follows: $\sigma_0=\sigma_1=1,\left(a_0, b_0, c_0\right)=(0.6,6,30)$, $\mathbf{a}_1 \sim \mathcal{N}\left(\mathbf{0}, 2 \cdot \mathbf{I}_{d}\right), \mathbf{b}_1 \sim \mathcal{N}\left(10 \cdot \mathbf{1}, 2 \cdot \mathbf{I}_{d}\right), \mathbf{c}_1 \sim \mathcal{N}\left(15 \cdot \mathbf{1}, 2 \cdot \mathbf{I}_{d}\right)$, where $\mathbf{1}$ is a vector of ones and $\mathbf{I}_{d}$ is an identity matrix, $d= 5$. 

\paragraph{Case 3.} We maintain all parameters consistent with \textbf{Case 2}, except for the sample size, which is defined as $N_k = 1000 - 200(k-1)$ for $k$ clients, where $k \in \{1, \ldots, 5\}$, resulting in imbalanced datasets.

\paragraph{Evaluation Metrics.} The parameters for each client vary, making direct comparison with the centralized parameters infeasible. Therefore, we exclude the centralized metric in this section and retain the other metrics as in \textbf{Case 1}.

The mean values and standard errors of the evaluation metrics for both balanced and imbalanced settings are reported in Table~\ref{tab:table 2} and Table ~\ref{tab:table 3}. 


\texttt{xFBCI} achieves the lowest A-RMSE and ATE error compared to the two benchmark models, showcasing its superior accuracy in parameter estimation and treatment effect evaluation. Notably, it maintains strong performance even in imbalanced cases, effectively handling variations in sample sizes across clients. This highlights its robustness and adaptability in heterogeneous data settings.

\begin{table}[H]
    \centering
\captionsetup{font=footnotesize}    
\begin{tabular}{c c c c}
\hline Models & A-RMSE & A-ATE & Error of ATE \\
\hline \texttt{Individual} & 0.225($\pm 0.020$)  & 20.3($\pm 1.65$) &0.82 \\
\hline \texttt{xFBCI} &$\bm{0.187(\pm 0.015)}$ &21.0($\pm 1.42$) & \bm{$0.15$}\\
\hline \texttt{Ditto} &0.204($\pm 0.015$) & 20.9($\pm 0.02$)&0.25 \\

\hline
\end{tabular}

\caption{Results of Case 2. We report the mean values and standard errors of 10 replications for each metric across the 3 benchmarks. The sample size of each client is: $N_k = 1000$, where $k \in\{1, \ldots, 5\}$.}

\label{tab:table 2}
\end{table}

\begin{table}[H]
    \centering
\captionsetup{font=footnotesize}    
\begin{tabular}{c c c c}
\hline Models & A-RMSE & A-ATE & Error of ATE \\
\hline \texttt{Individual} & 0.293($\pm 0.038$) &20.9($\pm 0.74$) & 0.89 \\
\hline \texttt{xFBCI} &\bm{$0.257(\pm 0.022)$} &21.07($\pm 0.44$) &\bm{$0.74$} \\
\hline \texttt{Ditto} & 0.347($\pm 0.013$) & 23.12($\pm 0.12$) & 1.32 \\

\hline
\end{tabular}

\caption{Results of Case 3. We report the mean values and standard errors of 10 replications for each metric across the 3 benchmarks. The sample size of $k$ client is: $N_k = 1000 - 200(k-1)$, where $k \in\{1, \ldots, 5\}$. }
\label{tab:table 3}
\end{table}

\subsection{Complex Heterogeneous Cases}


We explore even more challenging heterogeneous scenarios. Unlike the previous section, in this setting, the datasets for the clients are derived from different distributions. Furthermore, the dimensions, sample sizes, number of clients, and true parameters are different, introducing greater complexity and variability.

\paragraph{Case 4.} We set the number of clients to 10, where $k \in\{1, \ldots, 10\}$, $N_k = 300$, and $d = 5$. The 10 clients are divided into three groups, with the datasets and true parameters for each group simulated under distinct scenarios.  Specifically, the distribution of dataset for client 1 to 3 is $\mathrm{U}(-1,1)$, for client 4 to 6 is $\mathcal{N}(2,2)$, and for client 7 to 10 is $\mathcal{N}(4,2)$, each with their corresponding assigned true parameters. The dataset generation process is similar to above cases. The details are included in Table~\ref{tab: table case 4}.

\paragraph{Case 5.} We keep all parameters consistent with the \textbf{Case 4} except for the
sample size, which is determined by $N_k=300-20(k-1)$ for $k$ clients, where $k \in\{1, \ldots, 10\}$ (imbalanced dataset).

\begin{table}[H]
    \centering
    \captionsetup{font=footnotesize}  
    \small
\setlength\tabcolsep{3pt}

\resizebox{\textwidth}{!}{

\begin{tabular}{c c c c c c}
\hline Clients ($k$) & $X_k$ & $\left(a_0, b_0, c_0\right)$ & $\mathbf{a}_1$ & $\mathbf{b}_1$ & $\mathbf{c}_1$\\ 
\hline 1 - 3  & $\mathrm{U}(-1,1)$& $\left(0.5, 1.0, 2.0\right)$& $\mathcal{N}(\mathbf{0},2\cdot \mathbf{I}_{d})$ & $\mathcal{N}(6\cdot \mathbf{1},2\cdot \mathbf{I}_{d})$ & $\mathcal{N}(6\cdot \mathbf{1},2\cdot \mathbf{I}_{d})$\\
\hline 4 - 6  & $\mathcal{N}(2,2)$& $\left(-5.0, 2.0, 4.0\right)$& $\mathrm{Beta}(5\cdot \mathbf{1},1\cdot \mathbf{I}_{d})$ & $\mathcal{N}(3\cdot \mathbf{1},3\cdot \mathbf{I}_{d})$ & $\mathcal{N}(3\cdot \mathbf{1},3\cdot \mathbf{I}_{d})$\\
\hline 7 - 10 & $\mathcal{N}(4,2)$& $\left(-10, 6.0, 8.0\right)$& $\mathrm{Beta}(10\cdot \mathbf{1},5\cdot \mathbf{I}_{d})$ & $\mathcal{N}(5\cdot \mathbf{1},5\cdot \mathbf{I}_{d})$ & $\mathcal{N}(5\cdot \mathbf{1},5\cdot \mathbf{I}_{d})$\\
\hline
\end{tabular}}
\caption{Details of simulated dataset across $k$ clients, where $k \in\{1, \ldots, 10\}$. When \( k = 1, \ldots, 3 \), the dataset is sampled from \( \mathrm{U}(-1,1) \); when \( k = 4, \ldots, 6 \), it is sampled from \( \mathcal{N}(2,2) \); and when \( k = 7, \ldots, 10 \), it is sampled from \( \mathcal{N}(4,2) \), all with assigned parameters in the table.}
\label{tab: table case 4}
\end{table}

\paragraph{Evaluation Metrics.} In this section, the A-ATE is not representative, as the estimated ATE may vary significantly across different clients because of the various distributions of the datasets. Thus, we use the A-RMSE between estimated ATE $\tau_k$ of $k$ client with true ATE $\tau_{true}$ as a new metric to replace the A-ATE mentioned earlier. The A-RMSE of estimated ATE is computed as: 

$$
\text{A-RMSE (ATE)} = \sqrt{\frac{1}{K} \sum_{k=1}^{K} \left( \tau_k - \tau_{true} \right)^2}
$$

We report our results in Table~\ref{tab: table 5} and Table~\ref{tab: table 6}. 


We observe that as the dataset becomes increasingly heterogeneous across clients, the performance of all benchmarks declines, reflecting the challenges posed by greater variability in data distributions. Despite this, \texttt{xFBCI} consistently outperforms the other benchmarks, demonstrating its robustness and adaptability in handling heterogeneous scenarios. Its ability to deliver more accurate parameter estimation and treatment effect evaluation highlights its effectiveness, even in challenging settings with significant inter-client variability.

\begin{table}[H]
    \centering
\captionsetup{font=footnotesize}  
\begin{tabular}{c c c c}
\hline Models & A-RMSE ($\theta$) & A-RMSE (ATE)\\ 
\hline \texttt{Individual} & 0.602($\pm 0.124$)& 1.86($\pm 0.41$)\\
\hline \texttt{xFBCI} &\bm{$0.433(\pm 0.101)$} &\bm{$1.37(\pm 0.35)$}\\
\hline \texttt{Ditto} & 0.482($\pm 0.007)$& 1.49($\pm 0.45$) \\

\hline
\end{tabular}
\caption{Results of Case 4. We report the mean values and standard errors of 10 replications for each metric across the 3 benchmarks. The sample size of each client is: $N_k = 300$, where $k \in\{1, \ldots, 5\}$. }
\label{tab: table 5}
\end{table}

\begin{table}[H]
    \centering
\captionsetup{font=footnotesize}
\begin{tabular}{c c c c}
\hline Models & A-RMSE ($\theta$) & A-RMSE (ATE)\\ 
\hline \texttt{Individual} & 0.717($\pm 0.117$)& 2.32($\pm 0.21$)\\
\hline \texttt{xFBCI} &\bm{$0.492(\pm 0.164)$} &\bm{$1.51(\pm 0.12)$}\\
\hline \texttt{Ditto} &0.524($\pm 0.013$)  & 1.81($\pm 0.08$) \\

\hline
\end{tabular}
\caption{Results of Case 5. We report the mean values and standard errors of 10 replications for each metric across the 3 benchmarks. The sample size of each client is: $N_k=300-20(k-1)$, where $k \in\{1, \ldots, 10\}$. }
\label{tab: table 6}
\end{table}

\paragraph{Case 6.} We increase the number of clients to evaluate the performance of our method in the scenario with a larger client base. We set the number of clients to 20, where $k \in\{1, \ldots, 20\}$, $N_k = 300$, and $d = 10$. The distribution for client 1 to 10 is $\mathrm{U}(-1,1)$ and client 11 to 20 is $\mathcal{N}(0,1)$. This case can be viewed as the extension of \textbf{Case 2} with different number of clients  and variations in distributions, dimensions, sample size, and model parameters. The dataset generation process is similar to above cases. The details are included in Table~\ref{tab: table 7}.

\paragraph{Case 7.} We keep all parameters consistent with the \textbf{Case 6} except for the sample size, which is set as $N_k = 300 - 10(k-1)$ (imbalanced dataset).

\begin{table}[H]
    \centering
    \captionsetup{font=footnotesize}  
    \small
\setlength\tabcolsep{3pt}

\resizebox{\textwidth}{!}{
\begin{tabular}{c c c c c c}
\hline Clients ($k$) & $X_k$ & $\left(a_0, b_0, c_0\right)$ & $\mathbf{a}_1$ & $\mathbf{b}_1$ & $\mathbf{c}_1$\\ 
\hline 1 - 10  & $\mathrm{U}(-1,1)$& $\left(0.5, 1.0, 2.0\right)$& $\mathcal{N}(\mathbf{0},2\cdot \mathbf{I}_{d})$ & $\mathcal{N}(\mathbf{0},1\cdot \mathbf{I}_{d})$ & $\mathcal{N}(\mathbf{0},1\cdot \mathbf{I}_{d})$\\
\hline 11 - 20 & $\mathcal{N}(0,1)$& $\left(0.5, 1.0, 2.0\right)$& $\mathcal{N}(\mathbf{0},2\cdot \mathbf{I}_{d})$ & $\mathcal{N}(\mathbf{0},1\cdot \mathbf{I}_{d})$ & $\mathcal{N}(\mathbf{0},1\cdot \mathbf{I}_{d})$\\
\hline
\end{tabular}}
\caption{Details of simulated dataset across $k$ clients, where $k \in\{1, \ldots, 20\}$. When \( k = 1, \ldots, 10 \), the dataset is sampled from \( \mathrm{U}(-1,1) \); when \( k = 11, \ldots, 20 \), it is sampled from \( \mathcal{N}(0,1) \), all with assigned parameters in the table.  }
\label{tab: table 7}
\end{table}

We report the results in Tables \ref{tab: table 8} and \ref{tab: table 9}. It can be seen that the results in the balanced case are better than those in the imbalanced case, which is consistent with the results from above cases. Additionally, \texttt{xFBCI} consistently outperforms the other benchmarks in 20 clients, demonstrating its effectiveness in handling a larger client base. 

We also provide the box-plot of each baseline's result in Figure~\ref{fig:figure box}.

\begin{table}[H]
    \centering
    
\begin{tabular}{c c c c}
\hline Models & A-RMSE ($\theta$) & A-RMSE (ATE)\\ 
\hline \texttt{Individual} &$0.593(\pm 0.034)$ & 1.20($\pm 0.02$)\\
\hline \texttt{xFBCI} & \bm{$0.379(\pm 0.027)$} &\bm{$1.09(\pm 0.01)$}\\

\hline \texttt{Ditto} & 0.386($\pm 0.007$) & 1.10($\pm 0.01$) \\

\hline
\end{tabular}
\captionsetup{font=footnotesize}  
\caption{Results of Case 6. We report the mean values and standard errors of 10 replications for each metric across the 3 benchmarks. The sample size of each client is: $N_k = 300$, where $k \in\{1, \ldots, 20\}$. }
\label{tab: table 8}
\end{table}

\begin{table}[H]
    \centering
\captionsetup{font=footnotesize}  
    
\begin{tabular}{c c c c}
\hline Models & A-RMSE ($\theta$) & A-RMSE (ATE)\\ 
\hline \texttt{Individual} & 0.752($\pm 0.08$)& 1.27($\pm 0.15$)\\

\hline \texttt{xFBCI} & \bm{$0.423(\pm 0.051)$} &\bm{$1.12(\pm 0.13)$}\\

\hline \texttt{Ditto} &0.429($\pm 0.032$) & 1.15($\pm 0.02$) \\

\hline
\end{tabular}
\caption{Results of Case 7. We report the mean values and standard errors of 10 repeated rounds for each metric across the 3 benchmarks. The sample size of each client is: $N_k = 300 - 10(k-1)$, where $k \in\{1, \ldots, 20\}$. }
\label{tab: table 9}
\end{table}

\begin{figure}[H]
    \centering
   \captionsetup{font=footnotesize}
    \begin{subfigure}[b]{0.45\textwidth} 
        \centering
        \includegraphics[width=\textwidth]{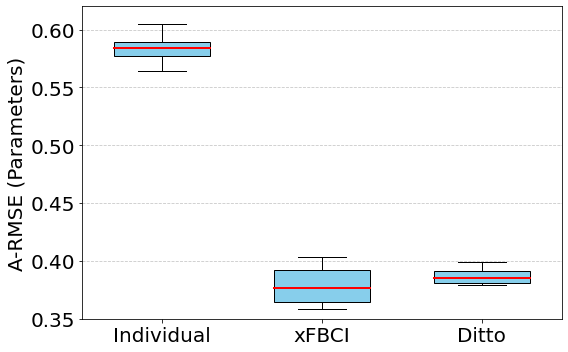}
        \caption{A-RMSE(Parameters)}
        \label{fig:dd}
    \end{subfigure}
    \hfill 
    \begin{subfigure}[b]{0.45\textwidth}
        \centering
        \includegraphics[width=\textwidth]{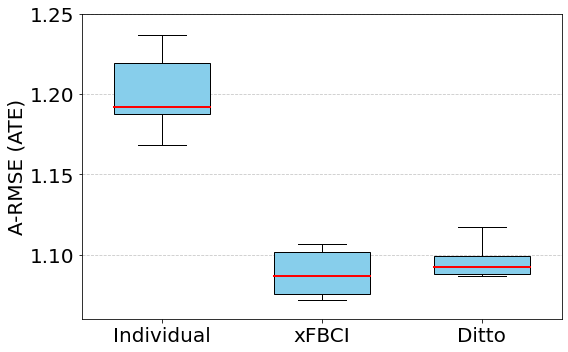}
        \caption{A-RMSE(ATE)}
    \end{subfigure}
    
    \label{fig:main}
     \caption{Box-plot of results in Case 6. }
\label{fig:figure box}
\end{figure}

\paragraph{Extreme Case.} In this case, we implement \texttt{xFBCI} in a scenario where the values in the dataset vary significantly and are substantially larger.  We set the number of clients to 10, where $k \in\{1, \ldots, 10\}$, $N_k = 300$, and $d = 10$. The dataset distribution for client 1 to 4 is $\mathrm{U}(0,30)$ and client 5 to 10 is $\mathcal{N}(2,2)$. The details are provided in Table~\ref{tab: table case e}.

The results for this extreme scenario are presented in Table~\ref{tab:tab10}. These results demonstrate that our proposed \texttt{xFBCI} outperforms the other benchmarks, even when the dataset range is exceptionally large across clients, showcasing its robustness in handling extreme variability.

\begin{table}[H]
    \centering
    \captionsetup{font=footnotesize}  
    \small
\setlength\tabcolsep{3pt}

\resizebox{\textwidth}{!}{

\begin{tabular}{c c c c c c}
\hline Clients ($k$) & $X_k$ & $\left(a_0, b_0, c_0\right)$ & $\mathbf{a}_1$ & $\mathbf{b}_1$ & $\mathbf{c}_1$\\ 
\hline 1 - 4  & $\mathrm{U}(0, 30)$& $\left(-2.0, 4.0, 6.0\right)$& $\mathcal{N}(\mathbf{0},3\cdot \mathbf{I}_{d})$ & $\mathcal{N}(4 \cdot \mathbf{1},3\cdot \mathbf{I}_{d})$ & $\mathcal{N}(4\cdot \mathbf{1},3\cdot \mathbf{I}_{d})$\\
\hline 5 - 10  & $\mathcal{N}(2,2)$& $\left(-2.0, 2.0, 4.0\right)$& $\mathrm{Beta}(10\cdot \mathbf{1},15\cdot \mathbf{I}_{d})$ & $\mathcal{N}(3\cdot \mathbf{1},3\cdot \mathbf{I}_{d})$ & $\mathcal{N}(3\cdot \mathbf{1},3 \cdot \mathbf{I}_{d})$\\

\hline
\end{tabular}}
\caption{Details of simulated dataset across $k$ clients, where $k \in\{1, \ldots, 10\}$. When \( k = 1, \ldots, 4 \), the dataset is sampled from \( \mathrm{U}(0,30) \); when \( k = 5, \ldots, 10 \), it is sampled from \( \mathcal{N}(2,2) \), all with assigned parameters in the table.}
\label{tab: table case e}
\end{table}

\begin{table}[H]
    \centering
\captionsetup{font=footnotesize}
\begin{tabular}{c c c c}
\hline Models & A-RMSE ($\theta$) & A-RMSE (ATE)\\ 
\hline \texttt{Individual} & 0.602($\pm 0.056$)& 1.48($\pm 0.45$)\\

\hline \texttt{xFBCI} & \bm{$0.353(\pm 0.107)$} &\bm{$1.37(\pm 0.23)$}\\

\hline \texttt{Ditto} &0.486($\pm 0.132$) & 1.47($\pm 0.02$) \\

\hline
\end{tabular}
\caption{Results of Extreme Case. We report the mean values and standard errors of 10 repeated rounds for each metric across the 3 benchmarks. The sample size of each client is: $N_k = 300$, where $k \in\{1, \ldots, 10\}$. }
\label{tab:tab10}
\end{table}

\section{Real Experiment} \label{sec:EHD}

To demonstrate the effectiveness of the proposed method, we apply \texttt{xFBCI} to real EHD printing datasets where the quality and specification of printed structures are influenced by multiple factors such as voltage, frequency, nozzle/stage moving speed, nozzle diameter, etc., as seen in Figure 4(a) and (b), the resulting output of different combinations of these factors impact the uniformity, resolution, and precision of the printed structure. When analyzing the effect of one factor (e.g., nozzle moving speed) on printing outcome, confounding variables (like voltage or frequency) can obscure the true relationship. Without controlling these confounders, relationships between input parameters and printed output may appear misleading. Moreover, the lack of precise control over confounders can result in inconsistent print quality and specification, such as irregular droplet sizes, inconsistent line widths, or poor adhesion.

\begin{figure}[H]
    \centering
   \captionsetup{font=footnotesize}
    \begin{subfigure}[b]{0.45\textwidth} 
        \centering
        \includegraphics[width=\textwidth]{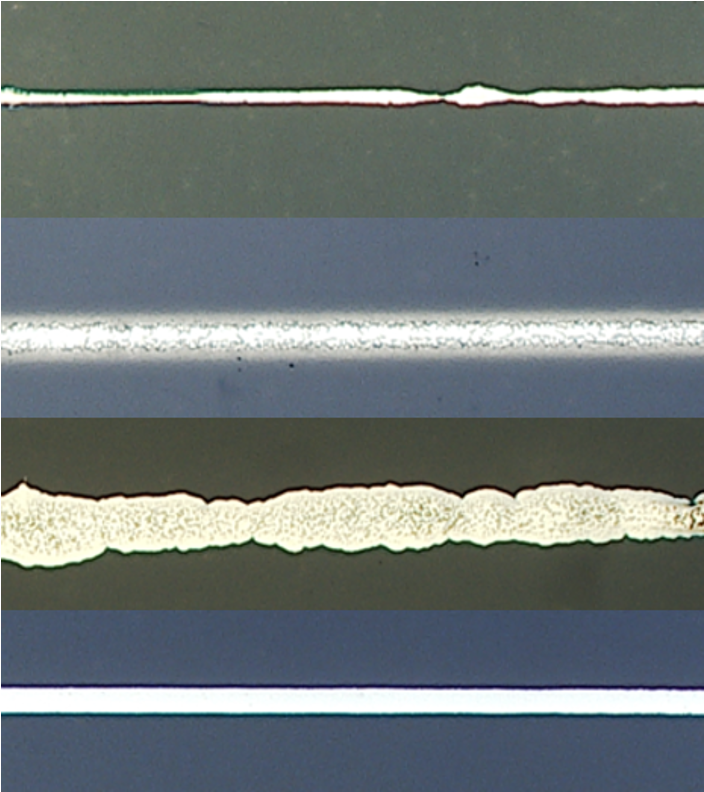}
        \caption{Lines with varying quality.}
        \label{fig:dd}
    \end{subfigure}
    \hfill 
    \begin{subfigure}[b]{0.45\textwidth}
        \centering
        \includegraphics[width=\textwidth]{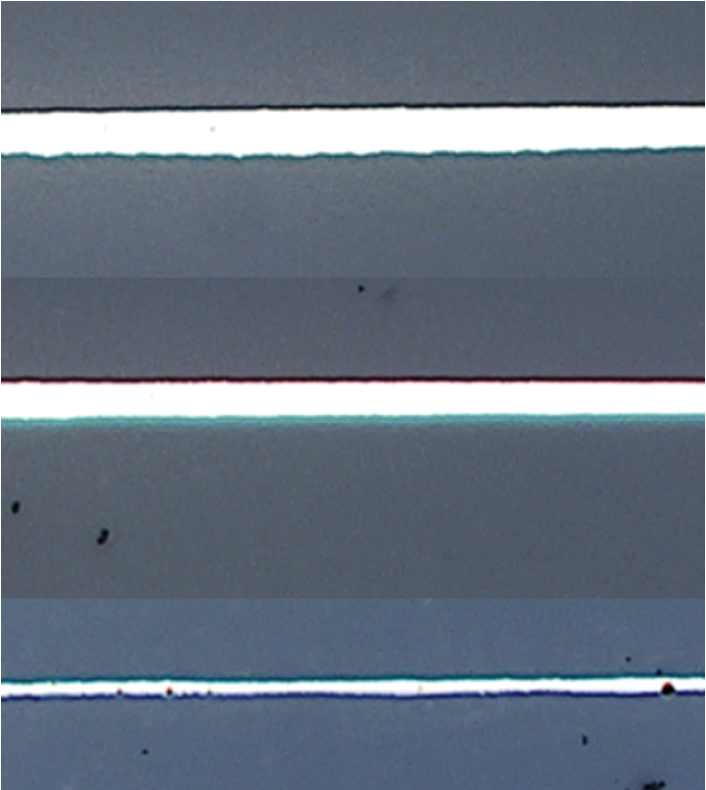}
        \caption{Lines with different specifications.}
    \end{subfigure}
    
    \label{fig:main}
     \caption{Examples of EHD printed lines.}
\label{fig:figure box}
\end{figure}


Propensity Score Matching balances the distribution of confounding variables between groups. For example, if we want to study the effect of high voltage versus low voltage on printing resolution, other factors vary across the data. It is challenging to match the exact experimental condition to study the effect of a single variable. PSM assigns a propensity score to each observation, representing the probability of receiving a specific treatment (e.g., high voltage), based on confounders. Observations with similar propensity scores are matched, ensuring that comparisons focus only on the effect of voltage, isolating confounding influences.


Our objective is to investigate the causal effects of all variables in the datasets collected from EHD printers using our proposed \texttt{xFBCI} method. Additionally, we compare the results obtained before and after applying PSM, demonstrating the enhanced performance and robustness of \texttt{xFBCI}.


\paragraph{Dataset.} We use a dataset from one EHD printer and divide it into two datasets each representing a separate ``Client" to utilize them in a manufacturing system that simulates a federated setting, where each dataset represents a local node contributing to the analysis without sharing raw data. The experimental variables considered in the study include: \textbf{Voltage, Frequency (Local), Duty Ratio, Nozzle/Stage Moving Speed, Stand-off Distance, Nozzle Diameter Size.} We use the \textbf{Printed Line Width} as a measurement of the printing specifications, where a narrower or more uniform line width indicates higher accuracy in material deposition, leading to better resolution and detail in the printed product. 

The dataset contains six variables that were generated randomly and independently for each data point to print two layers of 1-2 mm long silver lines on the top of a glass substrate using a commercial water-based ink known as JS-A191 Silver Nanoparticle Ink. After printing multiple lines, the substrate was placed on a heating plate and heated at 80°C for 5 minutes to dry them. Then, the slide was examined under a microscope, where the width of each line was measured three times and the average value was recorded. The dataset contains 210 data points. These six variables are interdependent and can influence one another, making them potential confounding factors in the analysis. We will explore the treatment effect of each variable separately. The final results comprise six treatment effect estimations and the corresponding mean squared error (MSE) values for each variable across both EHD datasets. When exploring the treatment effect of each variable, the respective variable is assigned a binary treatment value (0 or 1), and other variables are treated as covariates. 

\paragraph{Metrics.} We use the baselines \texttt{Ditto} and \texttt{Individual}, as described in Section \ref{sec:simulation}. However, the real dataset does not include the `true' parameters and ATE for direct comparison. Therefore, we evaluate the predictive performance of regression models on the two datasets, both before and after applying matching techniques in \texttt{xFBCI}, \texttt{Ditto}, and \texttt{Individual}. For simplicity, we focus on linear regression in this analysis. However, it is important to note that \texttt{xFBCI} is model-agnostic, meaning it is not dependent on the choice of the predictive model. To proceed, we first apply linear regression to the original datasets to estimate the predicted printed line width, $\hat{y}_{orignial}$. We then compare it to the actual printed line width, $y_{real}$.  Similarly, we estimate the printed line width, $\hat{y}_{after}$, using linear regression on the datasets processed with \texttt{xFBCI} (and other two baselines) and compare these predictions to $y_{real}$. By comparing the linear regression results before and after applying \texttt{xFBCI}, we can evaluate whether our proposed framework enhances the accuracy of estimated causal effects by effectively controlling for confounders and isolating the treatment's impact.

To evaluate and compare both sets of estimated results against the actual printed line width, we calculate the MSE, defined as:
$$
M S E=\frac{1}{N_k} \sum_{i=1}^{N_k}(\hat{y}_i-y_{real, i})^2
$$where $N_k$ is the sample size and MSE is calcualted on each client, $\hat{y}$ is the estimated values which could be replaced by $\hat{y}_{orignial}$ or $\hat{y}_{after}$ in this case. The smaller MSE represents the estimated values are more accurate.

\begin{figure}[H]
    \centering
    \captionsetup{font=footnotesize}
    \begin{subfigure}[b]{0.7\textwidth} 
        \centering
        \includegraphics[width=\textwidth]{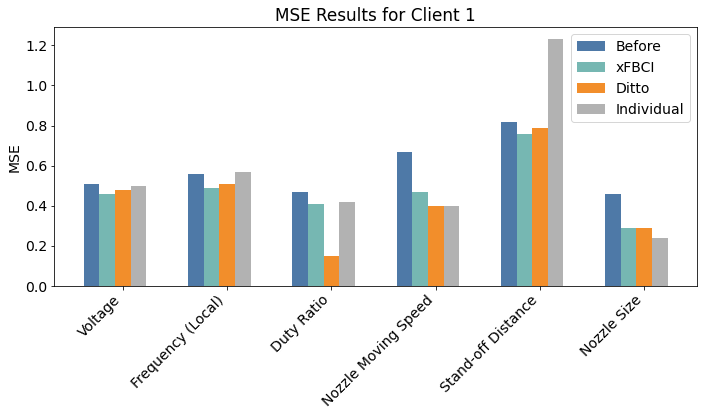}
        
        \label{fig:dd}
    \end{subfigure}
    \hfill
    \hspace{0.5cm} 
    \hfill
    \begin{subfigure}[b]{0.7\textwidth}
        \centering
        \includegraphics[width=\textwidth]{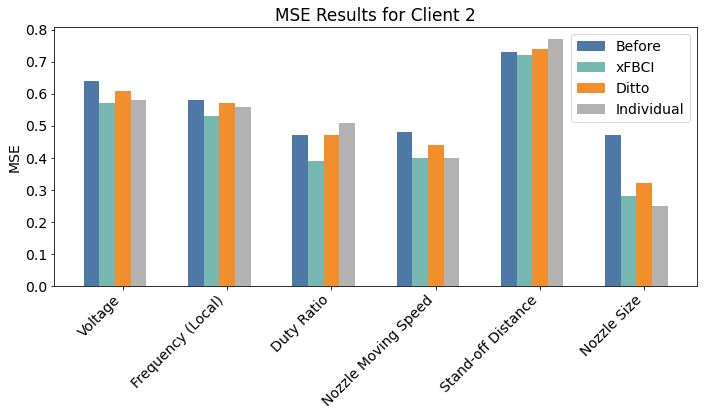}
        
    \end{subfigure}
    
    \label{fig:main}
    \caption{Bar-plot of the MSE results for before implementing matching (Before), after implementing matching of xFBCI, Ditto, and Individual for two clients. We utilize Client 1 and Client 2 here for representing the two EHD datasets.}
    \label{fig:figure 3}
\end{figure}

\paragraph{Results.} We report our results of estimated ATE in Table~\ref{table:11} and the MSE in Table~\ref{tab:12}. We also present MSE for a visual comparison of the estimation accuracy across different variables present in Figure~\ref{fig:figure 3} . 

The ATE quantifies the expected change in an outcome (i.e., the printed line width) in EHD printing, if the entire population were exposed to a treatment compared to not being exposed. For instance, an ATE of 27.6 for voltage indicates that, on average, applying the treatment (e.g., increasing or setting voltage to a specific level under study) results in a 27.6-unit increase in the printed line width. A positive ATE suggests a beneficial or enhancing effect of the treatment, while the magnitude provides a measure of its average impact. The results of the estimated ATE demonstrate that \texttt{xFBCI} yields results that are consistent with the fundamental principles of EHD printing. Although ATE values are different at each client, the signs are consistent. For example, faster nozzle moving would make the printed line width thinner since the material has less time to accumulate, and the treatment effect of it should be a negative value. The larger nozzle size causes a wider printed line width since it allows more material to be deposited at once, so the treatment effect is positive.

Understanding the ATE of key variables on printed line width can significantly advance the optimization of EHD printing processes. By quantifying the causal impact of these variables, manufacturers and researchers can make data-driven decisions to refine printing parameters for achieving desired line widths with greater precision and consistency. For instance, knowing that increasing voltage by a certain level results in a specific increase in line width allows for precise control over pattern formation, reducing trial-and-error experimentation. This insight can also inform the development of adaptive printing systems that dynamically adjust parameters in real-time, improving efficiency, reducing material waste, and enhancing scalability for industrial applications. Overall, the ATE provides a foundation for designing more reliable and efficient EHD printing workflows.

On the other hand, the MSE results in Table~\ref{tab:12} demonstrate that \texttt{xFBCI} achieves more accurate predictions, with smaller MSE values for 5 out of 6 variables after matching. While the other two benchmarks individually outperform \texttt{xFBCI} for a single variable, they fail to consistently reduce the MSE across all variables. The improved predictive accuracy of linear regression on line width after matching, compared to both pre-matching performance, has important implications for understanding and optimizing the EHD printing process. Matching helps reduce confounding by creating a more balanced dataset where treated and control groups have similar covariate distributions. This balance enhances the model's ability to isolate the true relationship between the variables of interest (e.g., voltage) and the printed line width, leading to more reliable predictions. The superior performance further highlights the value of incorporating causal inference techniques like propensity score matching to preprocess data, rather than relying solely on traditional regression approaches. This suggests that causal inference methods can serve as a powerful tool for improving predictive modeling in contexts where treatment effects and confounding factors significantly influence outcomes, ultimately leading to better-informed decisions in EHD printing optimization.


Overall, \texttt{xFBCI} provides more accurate line width predictions by effectively addressing the confounding variables in EHD printing. Furthermore, \texttt{xFBCI} enables researchers to identify potential confounders prior to conducting experiments, which can significantly reduce the costs associated with experimental design.



\begin{table}[H]

\centering
\captionsetup{font=footnotesize} 

\resizebox{\textwidth}{!}{
\begin{tabular}{lcccccc}
\toprule
\multirow{2}{*}{\textbf{Variable (Treatment)}} & \multicolumn{2}{c}{\textbf{xFBCI}} & \multicolumn{2}{c}{\textbf{Ditto}} & \multicolumn{2}{c}{\textbf{Individual}} \\
\cmidrule(lr){2-3} \cmidrule(lr){4-5} \cmidrule(lr){6-7}
 & Client 1 & Client 2 & Client 1 & Client 2 & Client 1 & Client 2 \\
\midrule
\textbf{Voltage}             & 27.6  & 16.3  & 26.7  & 19.4  & 18.4  & 30.8  \\
\textbf{Frequency (Local)}   & 7.6   & 3.9   & 15.5  & 7.5   & 21.9  & 2.3   \\
\textbf{Duty Ratio}          & -27.8 & -11.4 & -7.6  & -8.8  & -13.5 & -3.4  \\
\textbf{Nozzle Moving Speed} & -4.5  & -14.2 & -5.9  & -11.9 & -6.8  & -13.5 \\
\textbf{Stand-off Distance}  & 56.4  & 34.3  & 49.2  & 34.9  & 61.6  & 35.3  \\
\textbf{Nozzle Size}         & 25.0  & 24.5  & 25.0  & 23.2  & 25.0  & 23.2  \\
\bottomrule
\end{tabular}}
\caption{ATE results of xFBCI, Ditto, and Individual for two clients. Each variable is treated as binary value 0 and 1 when estimating its respective ATE. We utilize Client 1 and Client 2 here for representing the two EHD datasets.}
\label{table:11}
\end{table}

\vspace{1cm} 

\begin{table}[H]
\small
\setlength\tabcolsep{2pt}
\centering
\captionsetup{font=footnotesize}

\resizebox{\textwidth}{!}{
\begin{tabular}{lccccccccccccc}
\toprule
\multirow{2}{*}{\textbf{Variable (Treatment)}} & \multicolumn{4}{c}{\textbf{xFBCI}} & \multicolumn{4}{c}{\textbf{Ditto}} & \multicolumn{4}{c}{\textbf{Individual}} \\
\cmidrule(lr){2-5} \cmidrule(lr){6-9} \cmidrule(lr){10-13}
& \multicolumn{2}{c}{Client 1} & \multicolumn{2}{c}{Client 2} & \multicolumn{2}{c}{Client 1} & \multicolumn{2}{c}{Client 2} & 
\multicolumn{2}{c}{Client 1} & \multicolumn{2}{c}{Client 2} & \\
 & Before & After & Before & After & Before & After & Before & After & Before & After& Before & After \\
\midrule
\textbf{Voltage}             & 0.51 & \textbf{0.46} & 0.64 & \textbf{0.57} & 0.51 & 0.48 & 0.64 & 0.61 & 0.51 & 0.50& 0.64 & 0.58 \\
\textbf{Frequency (Local)}   & 0.56 & \textbf{0.49} & 0.58 & \textbf{0.53} & 0.56 & 0.51 & 0.58 & 0.57 & 0.56 & 0.57& 0.58 & 0.56 \\
\textbf{Duty Ratio}          & 0.47 & 0.41 & 0.47 & \textbf{0.39} & 0.47 & \textbf{0.15} & 0.47 & 0.64 & 0.47 & 0.42 & 0.47 & 0.51\\
\textbf{Nozzle Moving Speed} & 0.67 & \textbf{0.47} & 0.48 & \textbf{0.40} & 0.67 & 0.52 & 0.48 & 0.44 & 0.67 & 0.63& 0.48 & 0.40 \\
\textbf{Stand-off Distance}  & 0.82 & \textbf{0.76} & 0.73 & \textbf{0.72} & 0.82 & 0.79 & 0.73 & 0.74 & 0.82 & 1.23 & 0.73 & 0.77\\
\textbf{Nozzle Size}         & 0.46 & 0.29 & 0.47 & 0.28 & 0.46 & 0.29 & 0.47 & 0.32 & 0.46 & \textbf{0.24} & 0.47 & \textbf{0.25}\\
\bottomrule
\end{tabular}}
\caption{MSE results for xFBCI, Ditto, and Individual for two clients. Each variable is treated as binary value 0 and 1 when estimating its respective MSE. `Before' is the MSE of datasets before matching, and `After' represents the MSE after implementing the respestive method.  We utilize Client 1 and Client 2 here for representing the two EHD datasets. The smallest MSE among the 3 baselines for each variable is highlighted in \textbf{bold}.}
\label{tab:12}
\end{table}

\section{Discussion and Conclusion} \label{sec:con}
In this paper, we propose a framework for federated Bayesian causal inference, \texttt{xFBCI}. \texttt{xFBCI} offers strong scalability and explainability, excelling particularly in highly heterogeneous datasets. Simulation results demonstrate that our method outperforms individual approach and other state-of-the-art federated methods in causal effect estimating. Furthermore, our work addresses the existing gap in federated causal inference research within advanced manufacturing. A real-world application in EHD printing showcases the success of our framework. By leveraging propensity score methods to estimate the Average Treatment Effect and applying matching to enhance predictive accuracy, we provide a robust framework for understanding causal relationships and improving predictive modeling in optimizing EHD printing processes. Finally, we anticipate that reformulating this framework could extend its applicability to other domains. We leave these directions for future research.


\section{Acknowledgment} \label{sec:acknow}
The authors gratefully acknowledge support from the University of Wisconsin–Madison Fall Research Competition and Northeastern University, which was instrumental in completing this research successfully.








\bibliography{mybib}

\newpage
\appendix
\section{Appendix}
We provide the details of the hyperparameters used for the cases in Section~\ref{sec:simulation}. 

Table~\ref{tab:table13} summarizes the hyperparameters used in the implementation of \texttt{xFBCI}, including: \textbf{Communication Rounds (CR)}, \textbf{Learning Rate (LR)} for SGLD, \textbf{Step Size} for SGLD, \textbf{Burn-in} Sample Size, which refers to the number of discarded samples after collecting $\Theta$, and \textbf{Batch}, which denotes the proportion of $N_k$ used to determine the batch size. The batch size is obtained by taking the integer part of the product of the specified proportion and $N_k$. If $\textbf{Batch} \times N_k$ is not an integer, we round it down to its integer part. For example, if $N_k = 36$ and $\textbf{Batch} = 0.8$, then $36 \times 0.8 = 28.8$, and the resulting batch size is 28. \bm{$\alpha$} is the hyperparameter in scaled identity matrix $\alpha I$.

Table~\ref{tab:table14} summarizes the hyperparameters used in the implementation of \texttt{Ditto}, including: \textbf{Communication Rounds (CR)}, \textbf{Learning Rate for local devices (LRL)}, \textbf{Learning Rate for global server (LRG)}, step size for local devices and global server (\textbf{Step Size L}, \textbf{Step Size G}) in SGD.

\begin{table}[h!]
    \centering
    \label{tab:no_vertical_lines}
    \small
    \setlength\tabcolsep{2pt}
    \centering
    \captionsetup{font=footnotesize}

    \begin{tabular}{lcccccc} 
        \toprule
        \textbf{Case} & \textbf{CR} & \textbf{LR} & \textbf{Step Size} & \textbf{Burn-in}&\textbf{Batch} &\bm{$\alpha$}\\
        \midrule
        Case 1 & 20 & 0.05 & 700 & 100 &0.9&$1 \times 10^{-20}$\\
        Case 2, 3 & 30 & 0.001 & 700 & 100 &0.9&$1 \times 10^{-15}$\\
 
        Case 4, 5& 30 & 0.002  & 700 & 100&0.8 & $1 \times 10^{-20}$\\
         
        Case 6, 7 & 40&0.001 & 600 & 100&0.8 & $1 \times 10^{-20}$
        \\
        Case 8, 9 & 40&0.001 & 600 & 100&0.8 &$1 \times 10^{-15}$\\
        Extreme Case & 40&0.002 & 700 & 100&0.8&$1 \times 10^{-20}$   \\
        \bottomrule
        
    \end{tabular}
    \caption{xFBCI's hyperparameters used in Section~\ref{sec:simulation}. The hyperparameters include: \textbf{Communication Rounds (CR)}, \textbf{Learning Rate (LR)} for SGLD, \textbf{Step Size} for SGLD, \textbf{Burn-in} Sample Size, \textbf{Batch}, and \textbf{$\alpha$}. }
    \label{tab:table13}
\end{table}

\begin{table}[h!]
    \centering
    \small
    \setlength\tabcolsep{2pt}
    \centering
    \captionsetup{font=footnotesize}

    \begin{tabular}{lccccc} 
        \toprule
        \textbf{Case} & \textbf{CR} & \textbf{LRL} & \textbf{LRG}&\textbf{Step Size L} & \textbf{Step Size G}\\
        \midrule
        Case 1   & 20 & 0.005   & 0.005 & 500 &500\\
        Case 2   & 30 & 0.001  & 0.001 & 400 &400\\
        Case 3   & 30 & 0.002  & 0.001 & 400 &400\\
 
        Case 4 & 40 & 0.01  & 0.02 & 400 & 400 \\
        Case 5 & 40  & 0.01  & 0.01 &500 & 500  \\
        Case 6, 7 & 30  & 0.01  & 0.01&400 & 400  \\
        Extreme Case & 40 & 0.01 & 0.01&500 & 500 \\
        \bottomrule
    \end{tabular}
    \caption{Ditto's hyperparameters used in Section~\ref{sec:simulation}. The hyperparameters include: \textbf{Communication Rounds (CR)}, \textbf{Learning Rate for local devices (LRL)} in SGD, \textbf{Learning Rate for global server (LRG)} in SGD, step size for local devices and global server (\textbf{Step Size L}, \textbf{Step Size G}) in SGD.}
    \label{tab:table14}
\end{table}
\end{document}